\documentclass[conference,a4paper]{IEEEtran}
\IEEEoverridecommandlockouts
\usepackage{cite}
\usepackage{amsmath,amssymb,amsfonts}
\usepackage{algorithmic}
\usepackage{graphicx}
\usepackage{textcomp}
\usepackage{xcolor}
\def\BibTeX{{\rm B\kern-.05em{\sc i\kern-.025em b}\kern-.08em
		T\kern-.1667em\lower.7ex\hbox{E}\kern-.125emX}}

\usepackage{amsmath}
\usepackage{amsthm}
\usepackage{booktabs}
\usepackage{amsfonts}
\usepackage{amssymb}
\usepackage{enumerate}
\usepackage{enumitem}
\usepackage{wasysym}
\usepackage{mathrsfs}
\usepackage{graphicx, caption, subcaption}
\usepackage{wrapfig}
\usepackage{xcolor}

\usepackage[linesnumbered, ruled, noend]{algorithm2e}
\SetAlCapSkip{1em}
\SetKwInput{KwParam}{Params}
\SetKwInput{KwHyper}{Hyperparams}

\newcommand\mypseudocodecomment[1]{\textit{\color{blue} // #1 \\}}
\SetCommentSty{mycommfont}

\usepackage{xr}
\usepackage{hyperref}
\usepackage[nameinlink,capitalize]{cleveref}

\usepackage[margin=0.5cm,font=small,labelfont=bf,labelsep=endash,format=hang,labelformat=simple]{caption} [2022/02/20]
\numberwithin{figure}{section}

\newcommand{\R}{\mathbb{R}}

\newcommand{\N}{\mathbb{N}}

\renewcommand{\P}{\mathbb{P}}

\DeclareMathOperator{\ReLU}{ReLU}

\DeclareMathOperator{\softmax}{softmax}
\newcommand{\pluseq}{\mathrel{+}=}

\usepackage{chngcntr}
\usepackage{apptools}
\AtAppendix{\counterwithin{lemma}{section}}
\AtAppendix{\counterwithin{definition}{section}}
\AtAppendix{\counterwithin{theorem}{section}}

\newtheorem{problem}{Problem}

\newtheorem{theorem}{Theorem}

\begin{document}
	\title{Finding Hamiltonian cycles with graph neural networks\\
		\thanks{This work has been supported in part by the Croatian Science Foundation under the project Single genome and metagenome assembly (IP-2018-01-5886) and the Genome Institute of Singapore, A*STAR core funding}
	}
	
	\author{\IEEEauthorblockN{Filip Bosni\'c}
		\IEEEauthorblockA{\textit{Faculty of Electrical Engineering and Computing} \\
			\textit{University of Zagreb}\\
			Zagreb, Croatia\\
			ORCID 0000-0003-4888-5912}
		\and
		\IEEEauthorblockN{Mile \v{S}iki\'c}
		\IEEEauthorblockA{\textit{Laboratory of AI in Genomics} \\
			\textit{Genome Institute of Singapore, A*STAR}\\
			Singapore\\
			ORCID 0000-0002-8370-0891}
	}
	\maketitle
	\begin{abstract}
		We train a small message-passing graph neural network to predict Hamiltonian cycles on Erd\H{o}s-R\'enyi random graphs in a critical regime. It outperforms existing hand-crafted heuristics after about 2.5 hours of training on a single GPU. Our findings encourage an alternative approach to solving computationally demanding (NP-hard) problems arising in practice. Instead of devising a heuristic by hand, one can train it end-to-end using a neural network. This has several advantages. Firstly, it is relatively quick and requires little problem-specific knowledge. Secondly, the network can adjust to the distribution of training samples, improving the performance on the most relevant problem instances. The model is trained using supervised learning on artificially created problem instances; this training procedure does not use an existing solver to produce the supervised signal. Finally, the model generalizes well to larger graph sizes and retains reasonable performance even on graphs eight times the original size.
	\end{abstract}


	\section{Introduction} \label{sectionIntroduction}
	When dealing with problems that are computationally too costly to solve explicitly, such as NP-hard problems, it is common to rely on heuristics. The idea of using neural networks to train such heuristics is quite appealing and has attracted considerable interest over the years. One aims to enhance an algorithm, such as greedy search, with a neural network module that is trained to improve the decision-making of the algorithm. See \cite{machineLearningForCombinatorialOpt}, \cite{combinatorialOptimizationAndReasoning} or \cite{neuralAlgorithmicReasoningVelickovic} for an introduction and an overview of the area. In practice, problem instances typically come from a distribution with specific biases which are hard to describe explicitly. These can be exploited by a neural network.
	As an illustration, let us consider the Hamiltonian cycle problem (HCP), which is at the core of this paper (nodes in the \emph{cycle} can not repeat). It asks the following:
	\begin{problem}[HCP] \label{HCP}
		Determine whether or not there exists a cycle that passes through all vertices of a given graph. If it exists, such a cycle is called a \emph{Hamiltonian cycle}, and the graph is said to be \emph{Hamiltonian}.
	\end{problem}
	\noindent The general HCP is known to be NP-complete and thus computationally intractable. Currently, the fastest known exact solution algorithm is due to \cite{bjorklund2014determinant} and has worst-case complexity of $\mathcal{O}(1.657^n)$.
	
	As far as applications are concerned, HCP is used to improve runtimes of rendering engines, see \cite{arkin1996hamiltonTriangulations}. To do so, one solves the HCP for a dual graph of triangulation and renders the triangles in that order which reduces the number of points to process.
	Another application of HCP comes from genomics, more specifically, the problem of de novo genome assembly. The task here is to reconstruct the genetic material of an organism, i.e.\ the exact sequence of nucleobases on all of its chromosomes, from a large number of sequenced fragments called \emph{reads}. As chromosomes contain hundreds of millions bases, correctly assembling a single one is already a huge undertaking, see \cite{telomereToTelomerAssemblyCompleteHuman} for an example. Interpreting overlaps between reads as edges, after preprocessing and cleaning (see \cite{VrcekVelickovicSikicStepTowardsGenomeAssembly}), one ends up with a \emph{string graph} as proposed in \cite{myers2005fragment}. The Hamiltonian cycle in the string graph corresponds to the correct assembly of the chromosome. For more details see \cite{MihaiGenomeAssemblyReborn}, \cite{BakerDenNovoGenomeAssemlby}, \cite{VaserSikicRaven} and \cite{KolmogorovEtAlFly}.
	Both triangular meshes of 3d objects and string graphs of various assemblers (such as \cite{BakerDenNovoGenomeAssemlby} or \cite{VaserSikicRaven}) have specific structures and statistical properties arsing from the context. These could make solving the HCP easier but are difficult to exploit directly. We show here how to exploit them using graph neural networks in a similarly specific setting of Erd\H{o}s-R\'enyi random graphs.
	
	For HCP in general, heuristics based on Hopfield networks were already trained in the early 90-ties, see \cite{mehta1990neuralForHamilton,mehta1993analogNeuralForHamilton}. More recently, however, the area of geometric deep learning and graph neural networks has seen rapid developments and produced neural network layers such as message passing \cite{GilmerSchoenholzRileyVinyalisDahlMPGNN} or graph attention layers \cite{VelickovicGraphAttention}. These layers are built to exploit any graph structure in data and can handle arbitrarily large graphs with a limited set of parameters, resembling convolution layers in computer vision. They have found applications in image and text processing, combinatorial optimization, physics, chemistry \cite{GilmerSchoenholzRileyVinyalisDahlMPGNN} and biology \cite{MihaiGenomeAssemblyReborn}. See \cite{ZhouCuiHuEtAlGraphNeuralNetworksAReview} and \cite{BronsteinEtAlGeometricDeepLeraning} for a deeper dive into the area. In particular, they are excellent candidates for heuristics of graph-based problem.
	However, most efforts so far have been directed towards combinatorial optimization problems, the two-dimensional traveling salesman problem in particular. Heuristics for the 2d-TSP based on transformer architecture were trained in \cite{KoolHoofWellingAttentionTSP}, \cite{BressonTransformerTSP} and those based on graph neural networks in \cite{ZhihaoShikuiMonteCarloTreeSearchTSP} and \cite{JoshiLaurentBressonConvolutionalTSP}. The state-of-the-art result is achieved in \cite{BressonTransformerTSP} where a comprehensive list of references can be found as well.
	It has to be noted that previously mentioned models still perform worse than the Concorde TSP solver \cite{applegateConcordeSolver}, a state-of-the-art \emph{exact} solver based on branch and bound search combined with the cutting plane method. Nevertheless, theoretical complexities of neural network models are superior to Concorde. Let us also mention \cite{LearningCombinatorialOptimizationOverGraphs}, \cite{GNNMaximumConstraintSatisfcation} and \cite{OneModelAnyCSPGNN} which work with general combinatorial optimization and constraint satisfaction problems.
	
	In this paper we present a HCP solver based on \emph{graph} neural networks and show that it easily outperforms most hand-made heuristics. The code is available at \href{https://github.com/lbcb-sci/GNNs-Hamiltonian-cycles}{https://github.com/lbcb-sci/GNNs-Hamiltonian-cycles}.
	
	\section{Relation to TSP and 2d-TSP} \label{sectionConnectionToTSP}
	It is known that the HCP can be reformulated as a special case of the \emph{general traveling salesman problem (TSP)}:
	\begin{problem}[TSP]
		Given a graph with a non negative length assigned to each edge, find the shortest cycle passing through all its vertices.
	\end{problem}
	\noindent Hence, TSP solvers can be used for HCP and we shall exploit this by using \emph{Concorde TSP solver}, see \cite{applegateConcordeSolver}, to evaluate the performance of our models in \cref{sectionResultsDiscussion}. While it is tempting to assume that all papers studying TSP are immediately applicable to the HCP, this \emph{is not the case at all}. In particular, papers presenting neural network TSP solvers, such as \cite{BressonTransformerTSP,JoshiLaurentBressonConvolutionalTSP,KoolHoofWellingAttentionTSP} or \cite{ZhihaoShikuiMonteCarloTreeSearchTSP} only study the special case of \emph{two-dimensional TSP}:
	\begin{problem}[2d-TSP]
		Given a set of points in the unit square $[0, 1]^{2}$, find the shortest (in terms of Euclidean distance) cycle which passes through all of them.
	\end{problem}
	\noindent The 2d-TSP introduces two simplifications to the general TSP:
	\begin{itemize}
		\item graphs are always fully connected and
		\item distances between nodes comply with Euclidean structure (triangle inequality).
	\end{itemize}
	\noindent Only $2n$ point coordinates are required to describe a 2d-TSP instance, in contrast to $n^2 - n$ adjacency matrix weights needed for the general TSP. Moreover, 2d-TSP solvers can not be used to solve the HCP. On the contrary, we find it better to view the HCP and the 2d-TSP as two quite different aspects of the general TSP. The HCP focuses on complexities arising from discrete connectivity structure while the 2d-TSP deals with difficulties coming from the choice of edge lengths.
	
	\section{Problem setup} \label{sectionProblemSetup}
	We only consider simple, undirected graphs and denote a typical graph example by $G$ and its size (number of nodes) by $n$.
	The HCP is classically posed as a decision problem: \emph{Determine whether the graph contains a Hamiltonian cycle or not}. However, to put more emphasis on finding the actual cycle, which is important in practice, we also require that solvers produce at least one Hamiltonian cycle. In case the output of a solver is not a valid Hamiltonian cycle, which is straightforward to check, we assume the solver predicted that no Hamiltonian cycle exists.
	
	\subsection{Inputs and outputs}\label{sectionInputsOutputs}
	A solver receives as input a graph $G$ and outputs a walk $v_1 v_2 \ldots v_k$ on $G$ proposing a Hamiltonian cycle. The walk is considered to be closed if $v_1 = v_k$ and thus is a Hamiltonian cycle only if $k = n + 1$ and nodes $v_1, v_2, \ldots v_{k-1}$ are all distinct.
	
	\subsection{Evaluation distribution} \label{sectionCriticalRegime}
	The performance of HCP heuristics depends heavily on properties of graphs they are required to solve. Indeed, it is reasonable to have heuristics constructed specifically to achieve good performance on particular types of graphs, such as duals of triangulations or string graphs mentioned in \cref{sectionIntroduction}. As there are many possible applications of the HCP, finding a good class of evaluation graphs is a challenging task. Currently at least, there seems to be no agreed-upon class for this purpose. There are datasets of collected HCP problems, see, for example, \cite{tsplibPage} or \cite{FHCPdataset}, but they are not quite large enough to train neural networks on. A natural approach, used in early works such as \cite{mehta1990neuralForHamilton,mehta1993analogNeuralForHamilton,wagner1999antHeuristics} is to use random graphs generated by adding edges between pairs of vertices independently with a fixed probability $p \in (0, 1)$. Such random graphs are known as \emph{Erd\H{o}s-R\'enyi random graphs} with edge probability $p$. Papers working with 2d-TSP typically use a similar idea of evaluation on randomly generated problems, concretely the \emph{random uniform euclidean (RUE)} sets of two-dimensional points chosen uniformly at random from the unit square $[0, 1]^2$.
	
	However, using Erd\H{o}s-R\'enyi graphs with \emph{constant} edge probability parameter $p$ for evaluating the HCP has a major flaw. Intuitively it is clear that the HCP gets more difficult as the size of the graph increases. This is not the case for Erd\H{o}s-R\'enyi graphs with \emph{constant} $p$ as indicated by \cref{tableSupercritical}. It tracks performances of Concorde TSP solver and HybridHam heuristic from \cite{seeja2018hybridham}. The performance of either solver clearly improves as the graph size increases, suggesting that the problem is in fact getting easier. The issue is that large graphs end up having too many edges, leading to many Hamiltonian cycles thus making it easier to find one.
	
	\begin{table}[htb]
		\caption{Fraction of solved instances out of 5000 in supercritical regime, $p = 0.25$}
		\label{tableSupercritical}
		\centering
		\begin{tabular}{llllll}
			\multicolumn{6}{c}{} \\
			\toprule
			& \multicolumn{5}{c}{graph size} \\
			\cmidrule(r){2-6}
			Name     & $25$ & $50$ & $100$ & $150$ & $200$ \\
			\midrule
			Concorde & 0.80 & 1.0 & 1.0 & 1.0 &  1.0 \\
			HybridHam & 0.41 & 0.68 & 0.79 & 0.84 & 0.87 \\
		\end{tabular}
	\end{table}
	
	\noindent This can be mended by carefully decreasing parameter $p$ as the size of the graph increases. We rely on the following theorem from \cite{KomlosSzemerediLimitDistributionHamilton}.
	
	\begin{theorem}[Paraphrase of \cite{KomlosSzemerediLimitDistributionHamilton}, Theorem 1.]
		Let $\textnormal{ER}(n, p)$ denote the Erd\H{o}s-R\'enyi graph on $n$ nodes with edge probability parameter $p$. For every $p_H \in (0, 1)$ there is an explicit sequence $(p_n)_{n \in \N}$ such that
		\begin{equation*}
			\mathbb{P}\left(\textnormal{ER}(n , p_n) \text{ is Hamiltonian} \right) \xrightarrow{n \to \infty} p_H.
		\end{equation*}
		Concretely, one can take $p_n = \frac{\ln n + \ln \ln n - \ln \ln p_H^{-1}}{n-1}$.
	\end{theorem}

\begin{figure}[htb]
		\begin{subfigure}[b]{.48\linewidth}\includegraphics[width=\linewidth]{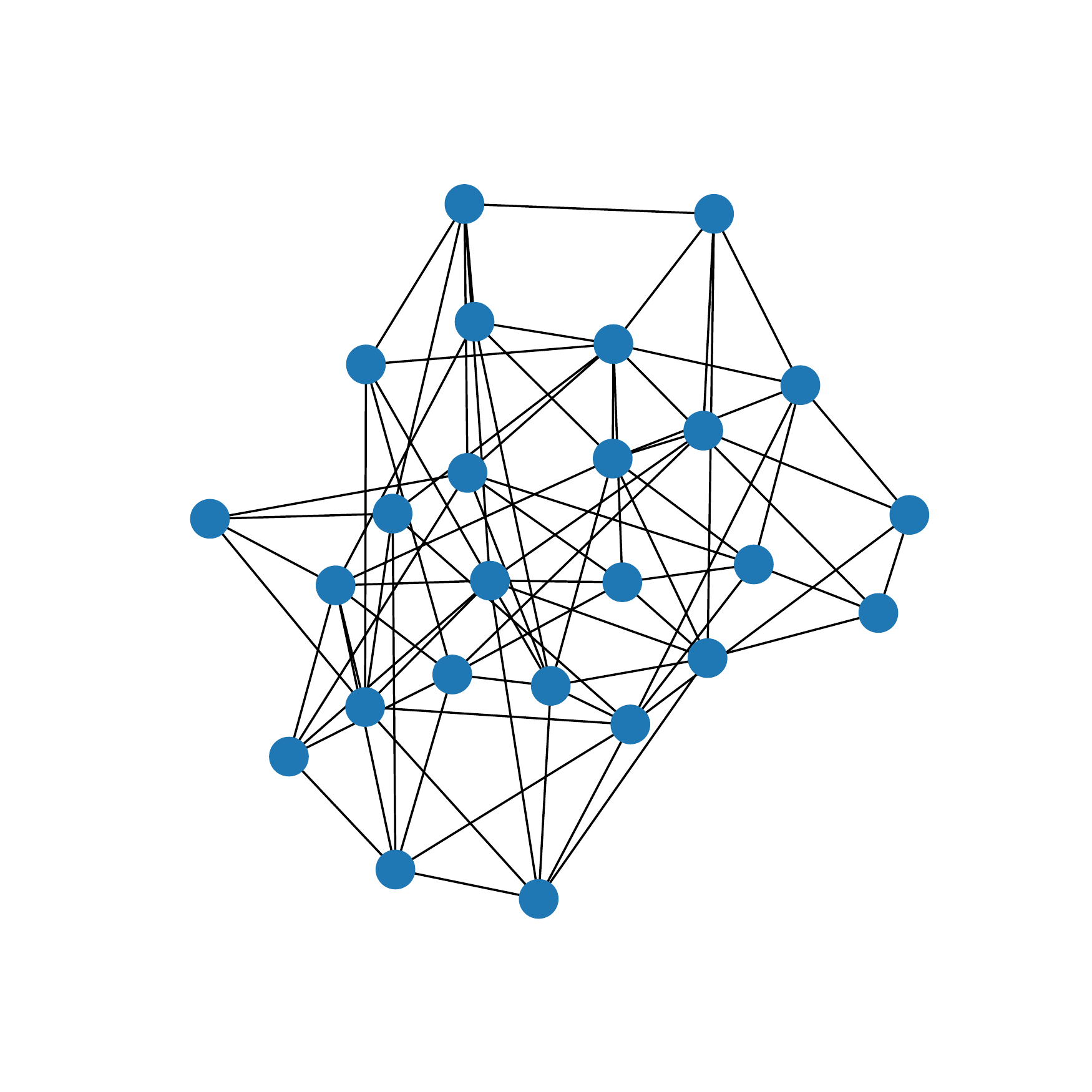}\caption{}\label{fig:1a}
		\end{subfigure}
		\begin{subfigure}[b]{.48\linewidth} \includegraphics[width=\linewidth]{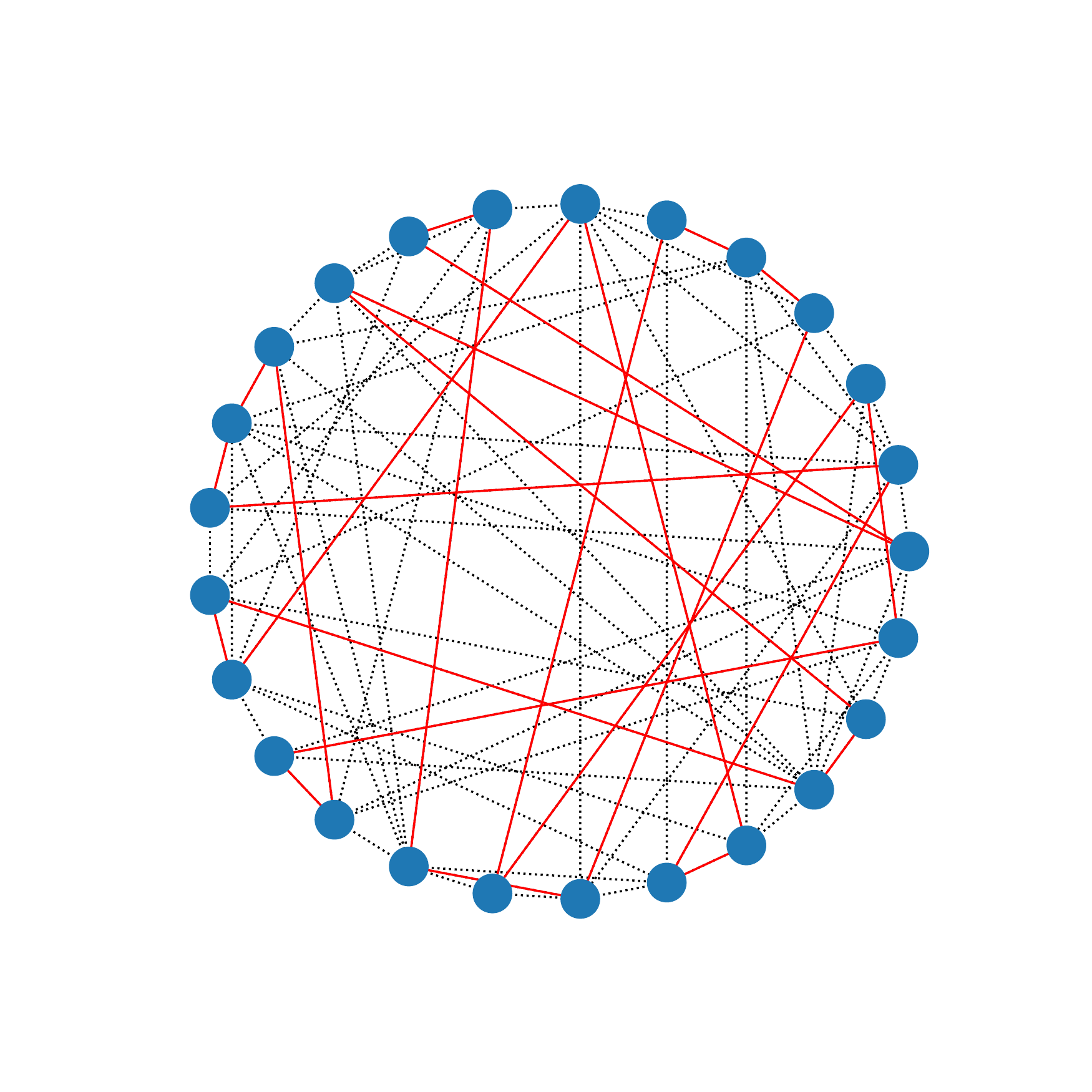} \caption{}\label{fig:1b}
		\end{subfigure}	
		
		\begin{subfigure}[b]{.48\linewidth}\includegraphics[width=\linewidth]{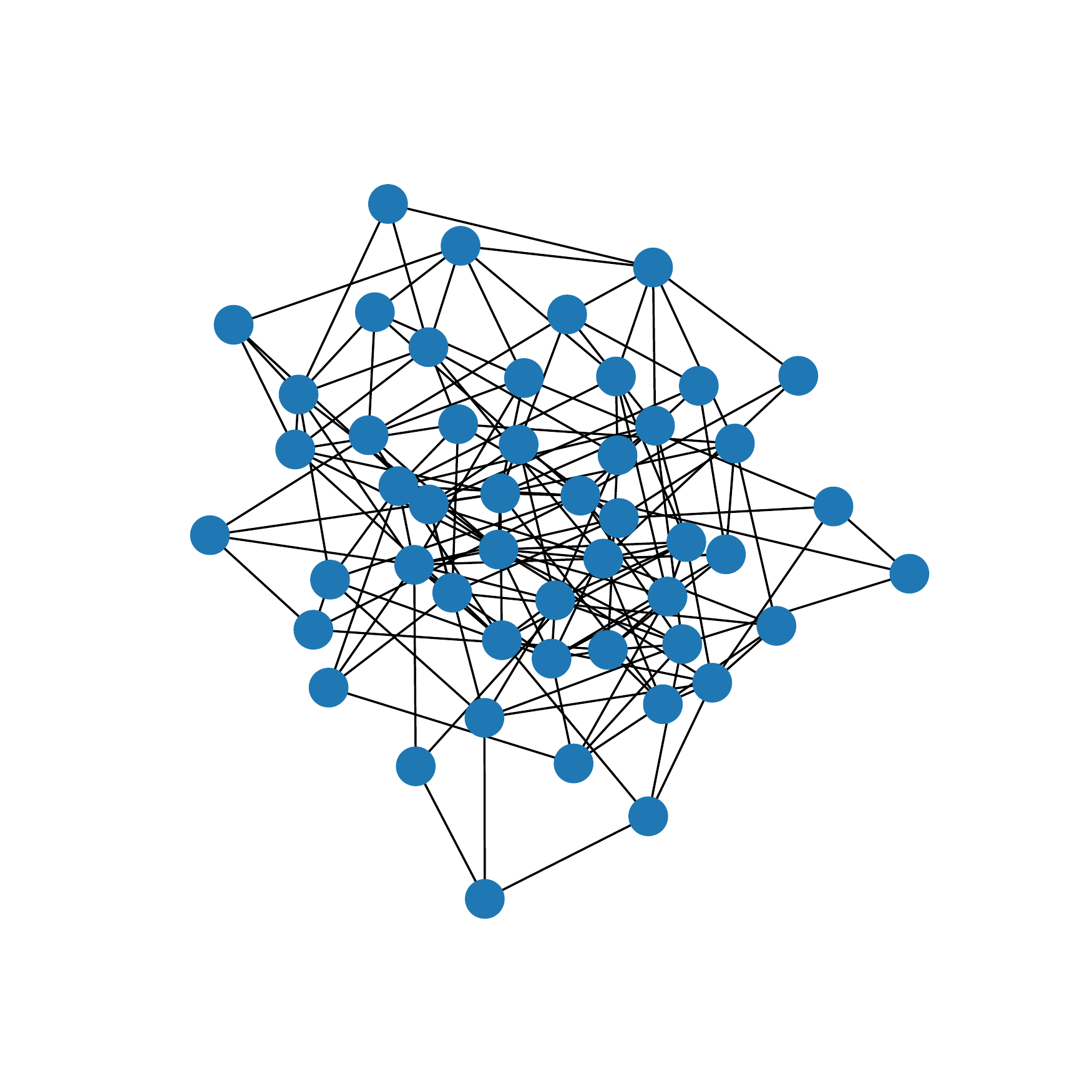}\caption{}\label{fig:2a}
		\end{subfigure}
		\begin{subfigure}[b]{.48\linewidth} \includegraphics[width=\linewidth]{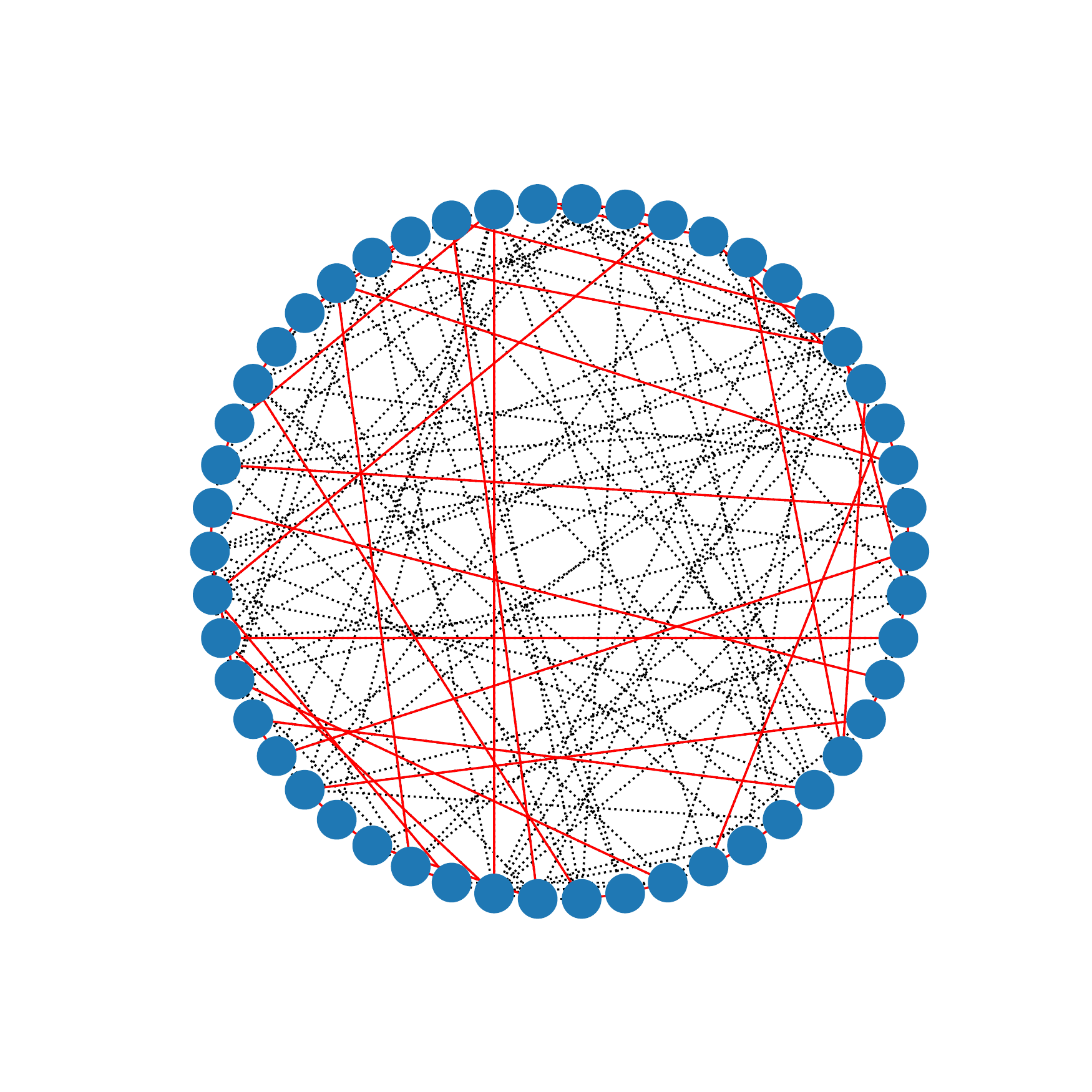} \caption{}\label{fig:2b}
		\end{subfigure}
		\caption{Examples of random graphs in the critical HCP regime. $25$ nodes in top and $50$ nodes in bottom row. Graphs in each row are identical. Right column graph is ordered in circle following a Concorde TSP solution, with the HP predicted by our basic model shown in solid red.}
		\label{fig1}
	\end{figure}

	\noindent In other words, for any $p_H$ there is a procedure of generating graphs such that they contain a Hamiltonian cycle with a probability approximately equal to $p_H$. We call this the \emph{critical regime} for the HCP. If the asymptotic behavior of $p_n$ is above the one from the previous theorem, we speak of \emph{the supercritical regime}. Examining the performance of Concorde solver in \cref{tableResults} shows that the empirical fraction of Hamiltonian cycles remains relatively stable and is fairly close to the asymptotic value of $p_H=0.8$.
	By controlling the existence probability of Hamiltonian cycles we control their expected number in a graph and hence also the difficulty of the HCP. This motivates our use of Erd\H{o}s-R\'enyi random graphs in the critical regime as the evaluation class. For simplicity, we use $p_H=0.8$ for the rest of the paper although other values of $p_H$ would work equally well. Two examples of random graphs in the critical regime are shown \cref{fig1}.
	
	\subsection{Datasets} \label{sectionDatasets}
	We work exclusively with generated datasets. Our test dataset is sampled from the evaluation distribution described in the previous \lcnamecref{sectionCriticalRegime} and consists of $5000$ Erd\H{o}s-R\'enyi graphs in critical regime with $p_H = 0.8$ for each size $n = 25$, $50$, $100$, $150$ and $200$. This sample size is large enough so that the fraction of Hamiltonian graphs stays within $\pm 2\%$ interval with $95\%$ probability for every size $n$. Train and validation datasets are generated from a different distribution described in \cref{sectionTraining}. They are never explicitly sampled. Instead, graph examples are generated on the fly when needed. The train dataset is \emph{limited} to graphs of size $25$ in order to emphasize generalization properties of the model.
	
	\section{Model details} \label{sectionModelDetails}
	
	Our model is autoregressive and decodes the Hamiltonian cycle a single node at a time. It begins by selecting a starting node and then chooses between neighbors in each following step. The chosen node is then appended to the partial solution and the process repeats until a node gets visited twice. There are two main components, a \emph{neural network component} that guides the neighbor selection at each step and a \emph{search algorithm} which combines selected nodes into a Hamiltonian cycle. Concretely, given a \emph{partial solution walk} $v_1 v_2 \ldots v_k$ at $k + 1$-th step of autoregressive decoding, the neural network component estimates with $\mathcal{P}(v \vert v_1 \ldots v_k)$ the probability that extending the walk by node $v$ will eventually lead to a Hamiltonian cycle (HC):
	\begin{align*} \label{equationExtensionProb}
		& \mathcal{P}(v \vert v_1 \ldots v_k) \approx \P \left(v_1 \ldots v_k v \subseteq \textnormal{HC} \big| v_1 \ldots v_k \subseteq \text{HC} \right).
	\end{align*}
	The search algorithm then selects the neighbor $v$ greedily according to estimated probabilities. It stops decoding when a node gets visited twice, i.e.\ $v \in \{v_1, \ldots v_k\}$, and returns $v_1 v_2 \ldots v_k v$ as the solution. The greedy approach is the simplest case of beam search algorithm with beam width $\beta = 1$. For beam width $\beta > 1$, at each step $k$ the algorithm keeps track of the top $\beta$ partial walks according to score
	\begin{align*}
		\mathcal{S}(v_1 v_2 \ldots v_k) &:= \prod_{j=1}^{k} \mathcal{P}(v_j \vert v_1 \ldots v_{j - 1}) \\ \nonumber
		&\approx \P(v_1 v_2 \ldots v_k \textnormal{ is contained in a HC})
	\end{align*}
	and extends them over all possible neighbors. A new set of top $\beta$ partial solutions is then selected and the process repeats. Clearly, larger beam width $\beta$ compensates for the performance of neural network at the cost of additional computations. While we report the performance of various beam widths in \cref{tableResults}, our basic model employs the simplest possible search algorithm ($\beta = 1$) in order to emphasize the neural network part.
	
	Our neural network uses \emph{persistent node features} $\mathbf{h}$ in the same way as in \cite{VelickovicYingPadovanoHadsellBlundellNeuralAlg}. These features are passed on between applications of the neural network, adding a sort of recurrent structure to the network. This provides a way for the network to access information from previous decoding steps.
	
	\subsection{GNN architecture}\label{sectionGNNArchitecture}
	Since \emph{graph neural networks (GNN)} form the central component of our model, HCP information needs to be represented in the suitable form. We represent the adjacency matrix of the graph as a list of edges and one hot-encode the following three node-level feature channels. Two channels to mark the start and the end node of the partial solution plus a channel to mark all nodes the solution contains. Note that this is precisely the information needed to correctly extend the walk by an unvisited node or close the HC if necessary.

	We employ the \emph{encode-process-decode} architecture analogue to the one used in \cite{VelickovicYingPadovanoHadsellBlundellNeuralAlg}. This means that our GNN is divided into the \emph{encoder}, \emph{processor} and \emph{decoder} networks. The whole GNN has around $22$ thousand parameters. Both encoder and decoder are single layer, fully connected networks with ReLU activation that operate on node features \emph{individually for each node}. The processor network, containing about $95\%$ of all parameters, is the core part. It is a residual stack of $5$ max-aggregation message passing layers, see \cite{GilmerSchoenholzRileyVinyalisDahlMPGNN} for more details. As names suggest, an input example is encoded, then processed and finally decoded by applying the above networks successively. In addition, we augment the output of the encoder with a randomized vector of features which was shown to improve the performance of GNNs in \cite{sato2021RandomFeaturesGNN}. \Cref{algorithmApplyNN} presents the pseudocode of a single forward pass. A "free" index $i \in G$ in a line indicates that this line should be repeated for each node; symbol $\bigoplus$ denotes concatenation in feature dimension; operator $\max_{j \sim i}$ stands for maximum over neighbors of $i$.
	
	\begin{algorithm}[h]
		\caption{$\texttt{ApplyGNN}(G, \mathbf{x}, \mathbf{h}; \theta)$.} \label{algorithmApplyNN}
		\DontPrintSemicolon
		\KwIn{$G$ - graph with $n$ vertices; \linebreak $\mathbf{x} \in \R^{(n, d_{\text{in}})}$ - partial walk repr.; \linebreak $\mathbf{h} \in \R^{(n, d_h)}$ - persistent features}
		\KwOut{$\mathbf{p} \in [0,1]^n$ - next-step probabilities per node.}
		\KwHyper{$d_{\text{in}} = 3, d_{h} = 32, d_{r} = 4, n_p = 5$}
		\KwParam{$\theta \equiv \{ W_E, b_E, W_P, b_P, \ldots\}$ - NN weights}
		\mypseudocodecomment{Encoder - Initialize features}
		$\mathbf{z}_i = W_E (\mathbf{x}_i \bigoplus \mathbf{h}_i) + b_E \in \R^{d_h - d_r}$ \label{lineEncoderInference} \\
		$\textbf{r} = \text{Uniform} \left([0, 1]^{n \times d_{r}} \right)$ $\in \R^{(n, d_{r})}$ \\
		$\mathbf{h}_i = \mathbf{z}_i \bigoplus \textbf{r}_i \in \R^{d_h}$\\
		\mypseudocodecomment{Processor - Apply residual max-MPNN layers}
		\For{$k = 1, 2, \ldots n_p$ \label{lineProcessorInference}}{
			$\mathbf{m}_i = \max_{j \sim i} \ReLU \left( W_M^{k} (\mathbf{h}_i \bigoplus\mathbf{h}_j \right)  + b_M) \in \R^{d_h}$ \\
			$\mathbf{h}_i = \mathbf{h}_i + \ReLU \left( W_P^{(k)} \left(\mathbf{h}_i \bigoplus \mathbf{m}_i \right) + b_P^{(k)} \right) \in \R^{d_h} $
		} \label{algorithmLineMessagePassing}
		\mypseudocodecomment{Decoder - Extract logits and probabilities}
		$\mathbf{l}_i = W_D (\mathbf{z}_i \bigoplus \mathbf{h}_i) + b_D \in \R$ \label{lineDecoderInference}\\
		\For{$i = 1, 2, \ldots, n$}{
			\If{$i \sim \textnormal{\texttt{GetLastNode}}(\mathbf{x}) $}{
				$\mathbf{l}_i = -\infty$ %\tcp{Mask out non-neighbors}
			}
		}
		$\mathbf{p} = \softmax \mathbf{l} \in \R^{n}$ \\
		\Return $\mathbf{p}$, $\mathbf{h}$
	\end{algorithm}
	
	\subsection{Training}\label{sectionTraining}
	Our supervised approach requires a large number of solved HCP instances during training. Even though they can easily be generated using existing HCP solvers, we will show it is possible to train on artificially generated graphs such that HCP solution is known in advance. We believe that such methods are useful when working with problems similar to HCP for which no exact solvers are available.
	The construction of a training example starts from a graph $G$ of arbitrary size but with no edges. A random permutation of nodes is then connected into a single cycle by adding appropriate edges into $G$. This will be a Hamiltonian cycle in the final graph and is stored as a supervision signal. Finally, for every pair of vertices in $G$ we add and edge connecting them with probability $p_{\textnormal{edge}} = 0.125$ (independently of other pairs). $p_{\textnormal{edge}}$ is treated as a training hyperparameter and was determined through experimentation.
	While the distribution of training samples generated in this way is quite different from the evaluation distribution which consists of ER graphs, the results show that the basic model still generalizes well. Note also that the final graph may have Hamiltonian cycles other than the original one. All such cycles are ignored during training.
	
	The training procedure is summarized in \cref{algorithmTraining}. A single training example consists of a graph $G$ and a Hamiltonian cycle $v_1 v_2 \ldots v_n v_1$ on $G$. The network is trained using \emph{teacher forcing} along this Hamiltonian cycle on the conditional cross-entropy loss $\mathcal{L}$ defined by
	\begin{equation*}
		\mathcal{L}\left(v_1 \ldots v_n v_1 \right) = - \sum_{i=2}^{n+1} \ln \left(\mathcal{P}(v_{i} \vert  v_1 \ldots v_{i - 1}) \right),
	\end{equation*}
	where $v_{n+1} := v_1$ for notational convenience.
	Remark that the summation index starts from $2$ because the choice of the first node in a cycle is completely arbitrary. Loss $\mathcal{L}$ is minimized over minibatches of 8 training examples using Adam optimizer with a learning rate of $10^{-4}$ for 2000 epochs of 100 gradient updates. The final model checkpoint was selected based on the fraction of solved instances on validation set generated in the same way as the training set. The whole training was performed on a single NVIDIA GeForce RTX 3080 GPU and took about 2.5 hours.
	Weight initialization and other optimizer hyperparameters are kept to default PyTorch 1.11.0 values, \cite{pyTorchPaszke2017automatic}.
	
	\begin{algorithm}[h]
		\caption{Training.} \label{algorithmTraining}
		\DontPrintSemicolon
		\KwIn{No input}
		\KwOut{$\theta_{\textnormal{final}}$ - trained parameters for the model.}
		\KwParam{$n = 25$ - training size; \linebreak $p_{\textnormal{edge}} = 0.125$ - generation edge probability; \linebreak maxStep = $20~000$ - nr. of gradient updates}
		$\theta = \texttt{RandomInitialization}()$ \\
		\For{$\textnormal{step} = 1, 2, \ldots \textnormal{maxStep}$}{
			G, c = \texttt{GenerateTrainExample}($n$, $p_{\textnormal{edge}}$) \\
			$\mathbf{h} = \texttt{GetInitialH}(\theta)$ \\
			$\textnormal{loss} = 0$ \\
			\For{$i = 1, \ldots, n + 1$}{
				$\mathbf{x} = \texttt{EncodeWalk}(c[:i])$ \\
				$\mathbf{p}, \mathbf{h} = \texttt{ApplyGNN}(G, \mathbf{x}, \mathbf{h}; \theta)$ \\
				\If{$i \neq 1$}{
					$\textnormal{loss} \pluseq  - \ln \mathbf{p}[c[i]]$
				}
			}
			$\theta = \texttt{GradientUpdate}(\nabla_{\theta} \textnormal{loss})$
		}
		\Return $\theta$
	\end{algorithm}
	
	\section{Results and discussion} \label{sectionResultsDiscussion}
	We evaluate the performance of our models by measuring the fraction of successfully solved problems on test dataset described in \cref{sectionProblemSetup} and compared it with following heuristics:
	\begin{enumerate}[label=(\roman*)]
		\item \emph{Concorde TSP solver} - the state-of-the-art exact TSP solver from \cite{applegateConcordeSolver},
		\item \emph{HybridHam} - an HCP heuristic from \cite{seeja2018hybridham},
		\item \emph{Ant-inspired heuristic} - an HCP heuristic presented in \cite{wagner1999antHeuristics},
		\item \emph{Least degree first heuristic} - simple greedy heuristic always selecting the neighbor with the lowest degree.
	\end{enumerate}
	Let us remark that the ant-inspired heuristic is a convergence procedure which we terminate after $5 n^2 \ln n$ steps. This bound matches the theoretical complexity of the basic model leading to a relatively fair comparison. In \cite{wagner1999antHeuristics}, authors suggest to terminate after $\mathcal{O}(n^3)$ iterations but this is very time consuming.
	We list evaluation results in \cref{tableResults} and average inference times in \cref{tableInference}. Keeping in mind that testing can be performed on a different sample of $5000$ graphs, the 95\% confidence interval for all values in \cref{tableResults} is below $\pm 0.02$. Models were run on a single NVIDIA GeForce RTX 3080 GPU while all other solvers were run on a single core of Intel Core i7-12700 processor. Note also that HybridHam, least degree first and ant-inspired heuristic were reimplemented in Python 3.8 and could be optimized for better performance.
	
	Our HCP setup makes it impossible for a solver to produce a false positive prediction. Consequently, all solvers have perfect precision and metrics such as $F_1$, $F_2$ are unnecessarily complicated. As the number of true positives (solvable HCPs) is stable by construction of the evaluation set (0.8 in the limit), accuracy, recall and fraction of solved instances have similar qualitative behavior. Thus we only report the fraction of solved instances for each model.
	
	\begin{table}[htb]
		\caption{Fraction of solved instance out of $5000$}
		\label{tableResults}
		\centering
		\begin{tabular}{llllll}
			\multicolumn{6}{c}{} \\
			\toprule
			& \multicolumn{5}{c}{graph size} \\
			\cmidrule(r){2-6}
			Name     & $25$ & $50$ & $100$ & $150$ & $200$ \\
			\midrule
			Concorde & 0.77 & 0.74 & 0.72 & 0.71 &  0.71 \\
			HybridHam & 0.39 & 0.52 & 0.35 & 0.23 & 0.15 \\
			Least deg.\ & 0.33 & 0.07 & 0.00 & 0.00 & 0.00 \\
			Ant-inspired & 0.73 & 0.09 & 0.00 & 0.00 & 0.00 \\
			\hline
			\textbf{Basic model} & 0.75 & 0.69 & 0.62 & 0.55 & 0.48 \\
			\hline
			Beam $\beta=2$ & 0.77 & 0.73 & 0.71 & 0.69 & 0.68 \\
			Beam $\beta=3$ & 0.77 & 0.73 & 0.72 & 0.70 & 0.70 \\
			Beam $\beta=5$ & 0.77 & 0.74 & 0.72 & 0.70 & 0.71 \\
			\bottomrule
		\end{tabular}
	\end{table}
	
	\begin{table}[hbt]
		\caption{Average inference time (ms) on $5000$ instance}
		\label{tableInference}
		\centering
		\begin{tabular}{llllll}
			\toprule
			& \multicolumn{5}{c}{graph size} \\
			\cmidrule(r){2-6}
			Name     & $25$ & $50$ & $100$ & $150$ & $200$ \\
			\midrule
			Concorde & 27.4 & 29.2 & 35.0 & 44.3 & 56.3 \\
			HybridHam & 1.1 & 2.5 & 6.0 & 10.2 & 15.3 \\
			Least deg. & 0.6 & 1.3 & 2.9 & 4.8 & 6.6 \\
			Ant-inspired & 9.9 & 56.0 & 289 & 755 & 1460 \\
			\hline
			\textbf{Basic model} & 29.4 & 61.3 & 130 & 209 & 293 \\
			\hline
			Beam $\beta=2$ & 58.5 & 151 & 448 & 933 & 1514 \\
			Beam $\beta=3$ & 87.8 & 227 & 673 & 1410 & 2284 \\
			Beam $\beta=5$ & 146 & 379 & 1129 & 2376 & 3845 \\
			\bottomrule
		\end{tabular}
	\end{table}	
	In conclusion, after only a few hours of training our basic model clearly outperformed existing heuristic solvers without using any pre-solved HCP. We believe that techniques similar to the ones presented here can be used to quickly develop heuristic for variations or generalizations of the HCP. For example, the task of finding the longest cycle in a graph. Or the task of finding the route of minimal length which covers all the nodes in the graph (some of them maybe more than once). The class of Erd\H{o}s-R\'enyi random graphs is used for simplicity and evaluation convenience since it allows for rough estimate of the difficulty of the HCP with respect to its size. Another class of graphs can be used just as well, provided that it is specific enough so that the neural network can exploit its statistical or structural peculiarities. But this typical happens with graph instances coming from practical problems. Moreover, polynomial complexity of $\mathcal{O}(n^2 \log n)$ for our basic model is superior to exponential complexity of exact solvers. For example, Concorde TSP solver on the RUE 2d-TSP instances was experimentally found to have complexity of $\mathcal{O}(1.24^{\sqrt{n}})$ in \cite{hoos2014empiricalComplexityConcorde}, although it is not clear how this translates to the critical regime HCP.
	Nevertheless, neural network solvers are yet to achieve reasonable performance on large input graphs and Concorde TSP solver remains the best-performing HCP solver. This comes as no surprise since Concorde also outperforms all existing neural network solvers for the 2d-TSP problem.
	
	\section{Ablation study \& training stability} \label{sectionAblationAndStability}
\begin{figure}[htb]
		\includegraphics[width=\linewidth]{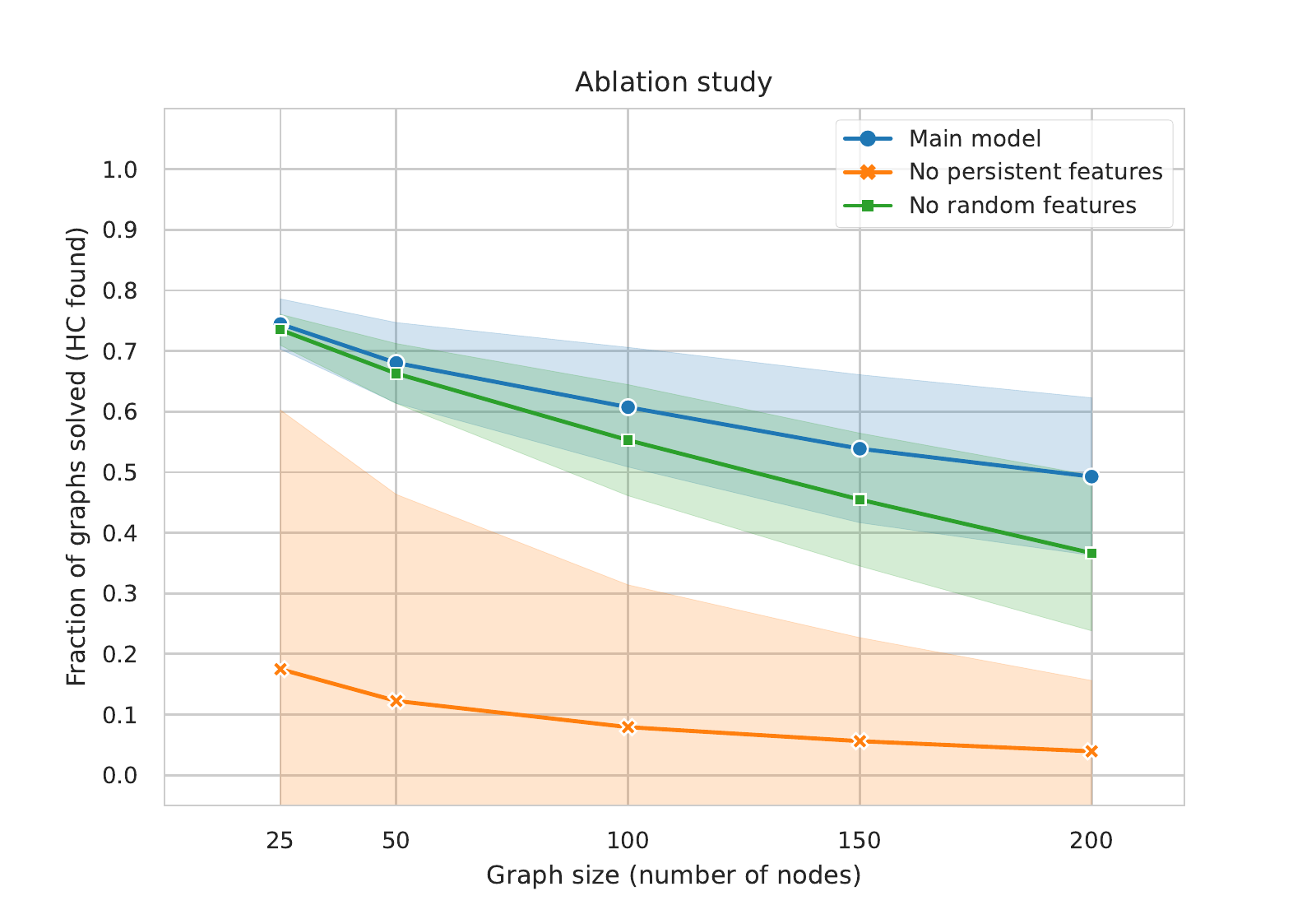}
		\caption{Average performance of models retrained 5 times. Shaded regions indicate intervals of 2 standard deviations} \label{figAblation}
	\end{figure}

	The neural network component from \cref{sectionModelDetails} is enhanced with persistent features and vectors of randomized features but can function without either of them. To estimate their importance, we separately removed each one and trained the corresponding reduced model 5 times from scratch. Average performances and confidence intervals of 2 standard deviation are shown in \cref{figAblation}.

	\noindent As shown on \cref{figAblation}, persistent features play a crucial role in our model. Without them the model can fail to converge during training. This is probably because persistent features allow the model to updated its internal node representations throughout decoding process which results in an RNN-like behavior and consequently increases the range of message passing neural network layers. The use of randomized features is not as significant but becomes noticeable when generalizing to large graphs. Note also that \cref{figAblation} shows the standard deviation of training procedure for the main model to be around 5\% of graphs solved.


	\bibliographystyle{plain}
	\bibliography{references}
	
\end{document}

\maketitle
	\appendix
	\section{Appendix}\label{appendix}	
	\setcounter{algocf}{0}
	\renewcommand{\thealgocf}{A.\arabic{algocf}}
	\setcounter{table}{0}
	\renewcommand{\thetable}{A\arabic{table}}
	
	\subsection{Graph theory basics} \label{sectionGraphNotation}
	For the sake of completeness, we introduce some basic notions from graph theory used in this paper. A \emph{graph} $G = (V, E)$ consists of a set $V$ of \emph{vertices}, a.k.a.\ \emph{nodes}, and a set $E \subset V \times V$ of \emph{edges}. An edge $(v,w) \in E$ represents an abstract connection from node $v$ to node $w$. The \emph{size} of graph $G$ is defined as the number of nodes it contains. A graph is \emph{undirected} if $(v, w) \in E$ implies $(v, w) \in E$, otherwise it is \emph{directed}. On an undirected graph $G$, we say that nodes $v, w \in G$ are neighbors if they are connected by an edge, i.e.\ $ (v, w) \in E$. This is denoted by $v \sim w$. A graph is \emph{simple} if it has no self-loops, that is, there is no node $v \in V$ such that $(v, v) \in E$. As a slight abuse of notation, the statement "$v \in G$" is considered equivalent to "$v \in V$". 
	
	A \emph{walk} on the graph $G$ is a sequence of nodes $v_1 v_2 \ldots v_k$ in $G$, such that every two successive nodes are neighbors. The length of a walk is equal to the number of nodes it contains. A \emph{path} is a walk in which all nodes are distinct, and a \emph{cycle} is a walk in which the first and the last nodes are the same, and all the other nodes are distinct.
	Assuming $V = \{1, \ldots, n\}$ (which can always be accomplished by relabeling nodes), the adjacency matrix of an undirected graph $G$, denoted by $\mathscr{A}$, is an $n \times n$, zero-one, a symmetric matrix such that $\mathscr{A}_{ij} = 1$ if and only if $i$ and $j$ are neighbors. A graph $G$ can be completely reconstructed from $\mathscr{A}$.
	In this paper, only simple, undirected graphs are considered.
	
	\subsection{Formal statements and proofs} \label{appendixSectionTSP}
	In this section, we list several simple statements concerning HCP and related problems.  
	\begin{lemma} \label{hamiltonDifficultWithSizeLemma}
		Let $HCP_n$ denote the Hamiltonian cycle problem posed for graphs of size exactly $n$, $n \geq 2$, and choose arbitrary $k \geq 2$. Any exact solver that can solve $HCP_{n+k}$ can also solve $HCP_n$ with time and space complexities changing up to a term of order $\mathcal{O}(n+k)$.
	\end{lemma}
	\begin{proof}
		Let us fix $n, k \in \N$, $n, k \geq 2$ and assume there is a solver that can solve $HCP_{n+k}$. Let also $G$ be an arbitrary graph with $n$ nodes. Choose an arbitrary node $v \in G$ and define graph $G'$ as follows. The node set of $G'$ consists of all nodes from $G$ except $v$, name them $v_1, v_2, \ldots v_{n-1}$, and $k + 1$ additional nodes $w_1, \ldots w_{k+1}$. The set of edges in $G'$ consists of:
		\begin{itemize}
			\item all edges from $G$ between vertices $v_1, \ldots v_{n-1}$,
			\item edges $(w_{i-1}, w_{i})$ and $(w_i, w_{i+1})$ for $i = 2, \ldots k$ (notice that $k \geq 2$ was assumed) and
			\item edges $(x, w_1), (w_{k+1}, x)$ for all nodes $x \in G$ such that $x$ and $v$ are neighbors in $G$.
		\end{itemize}
		We symmetrize this edge set so that $G'$ is also an undirected graph. Clearly, $G'$ has $n + k$ vertices and can therefore be solved with the solver selected in the beginning. Suppose first that the solver finds a Hamiltonian cycle $c'$ in $G'$. By definition of $G'$, the only neighbors of $w_i$ for $i=2, \ldots k$ are $w_{i-1}$ and $w_{i+1}$ and thus the cycle $c'$ must contain either a walk $w_1 w_2 \ldots w_{k+1}$ or $w_{k+1} w_{k} \ldots w_1$. Replacing this part of $c'$ with a walk consisting only of node $v$ gives a Hamiltonian cycle $c$ on the original graph $G$. Indeed, $c$ is a valid cycle because $w_1$ and $w_{k+1}$ are defined to have the same neighbors as $v$, and it must contain all nodes $v_1, \ldots v_{n-1}$ as they are all contained in $G'$. On the other hand, in case the solver does not find a Hamiltonian cycle in $G'$, we claim that there is no Hamiltonian cycle in $G$. Suppose the opposite, that $c$ is some Hamiltonian cycle in $G$. Replacing the node $v$ in this cycle with a walk $w_1 w_2, \ldots w_{k+1}$ gives a Hamiltonian cycle $c'$ on $G'$. Indeed, by definition, $w_1$ and $w_k$ are connected to all neighbors of $v$, and $c'$ contains all nodes in $G'$ leading to the contradiction. Note also that procedure described above basically consists of adding $k + 1$ nodes and less than $2n + 2k$ edges. Thus the time and space complexity change by a term of order $\mathcal{O}(n+k)$.
	\end{proof}
	
	\begin{lemma} \label{lemmaHCPisTSP}
		Let $G$ be an arbitrary graph and denote by $n$ its size. For a fully connected graph $G' = (V, E')$, $E' = V \times V$, there exists an assignment of lengths $l : E' \to [0, 1]$ such that the following is true. A solution of HCP on $G$ exists if and only if the solution of TSP on $G'$ has length $n/2$. Furthermore, these solutions coincide.
	\end{lemma}
	\begin{proof}
		Simply define length $l : E' \to [0, 1]$ as
		\begin{equation*}
			l(e) = \begin{cases}
				\frac{1}{2} \textnormal{ if } e \in E, \\
				1 \textnormal{ otherwise.}
			\end{cases}
		\end{equation*}
		Now, if cycle $c$ is a solution of HCP on $G$, then the length of $c$ (i.e.\ the sum of all edges in $c$) is $n/2$. From the definition of $l$, it is clear that no cycle with $n$ edges can have a length less than $n/2$. Thus $c$ is the shortest path in $G'$ and hence the solution of TSP. The other way around, if $c$ is the shortest cycle in $G'$ of length $n/2$, then by definition of $l$, all edges in $c$ are also in $E$. Thus, $c$ is a Hamiltonian cycle, a solution of HCP. Note that solutions coincide in either case.
	\end{proof}
	
	\subsection{Random graphs} \label{appendixSectionRandomGraphs}
	Let us start by defining the Erd\H{o}s-R\'eny random graph. This is one of the simplest models of random graphs, and it has been studied extensively.
	\begin{definition}
		An \emph{Erd\H{o}s-R\'eny random graph} on $n$ nodes, denoted by $G(n, p)$, is constructed by adding random edges to the graph of $n$ isolated nodes. For each pair of nodes, an edge connecting them is added with probability $p \in [0, 1]$ independently from all other pairs. In other words, \emph{Erd\H{o}s-R\'eny model} is a probability distribution on the set of all graphs with $n$ edges defined by the density
		$$ p(G) = p^{\vert E \vert} (1-p)^{\binom{n}{2} - \vert E \vert}$$
		where $E$ denotes the edge set of $G$. An \emph{Erd\H{o}s-R\'eny random graph} $G(n, p)$ is then simply a random variable with the above distribution.
	\end{definition}
	
	As far as 2d-TSP is concerned, it is common to use the following set of random 2d points.
	\begin{definition} \label{defninitionRUE}
		\emph{Random uniform euclidean (RUE)} set of points is a collection of points in $[0, 1]^2$ chosen independently according to the uniform distribution on the square $[0, 1]^2$.
	\end{definition}
	
	\subsection{Error estimates} \label{appendixSectionErrorEstimates}
	Measuring the accuracy of HCP solvers requires a large number of problem instances to keep the variance of results to a reasonable level. Let us suppose we work with a fixed heuristic solver. An experiment during evaluation consists of generating a single Erd\H{o}s-R\'eny random graph, solving it with the solver and either obtaining a Hamiltonian cycle or not. Such an experiment can be represented by a Bernoulli random variable $X_1 \in \{0, 1\}$ with $1$ indicating that a Hamiltonian cycle was found. The empirical fraction of solved problems, measured after $n$ experiments, is then given by 
	$$ a_n = \frac{\sum_{i=1}^{n} X_i}{n}.$$
	As multiple experiments are independent and identically distributed, $\sum_{i=1}^{n} X_i$ is a binomial random variable with a tail that can be estimated using Chernoff bound from \cite{mcdiarmid1998concentration}
	\begin{equation*}
		\mathbb{P}\left( \left\vert \sum_{i=1}^{n} X_i - n \mathbb{E} X_1 \right \vert \geq n \varepsilon \right) \leq 2 e^{-2 n \varepsilon^2} \qquad \forall \varepsilon > 0.
	\end{equation*}
	Hence, for a fixed error threshold of $\varepsilon$ and confidence interval of $1 - p$ one needs
	\begin{equation*}
		n \geq \frac{\varepsilon^{-2} \ln(2/p)}{2}
	\end{equation*}
	to make sure that the above tail bound holds. Specifically, taking $\varepsilon = 0.02$ and $p=0.05$ requires $n \geq \frac{1}{2} \cdot 50^2 \cdot \ln(40) = 4611.10$ graphs. This is why our evaluation is performed on $5000$ graphs.
	
	\subsection{Ablation study details}
	The following table lists performances of all models trained during ablation study.
	
	\begin{table}[htb]
		\centering
		\begin{tabular}{llllll}
			\toprule
		\end{tabular}
	\end{table}	
	\begin{table}[htb] \caption{Average performance on $5000$ graph instance} \label{appendixTableAblation}
		\begin{tabular}{l|r|r|r|r|r}\hline%
			 & \multicolumn{5}{c}{graph size} \\\hline
			Model name & 25 & 50 & 100 & 150 & 200 \\\hline
			\begin{filecontents*}{ablations.csv}
name,25,50,100,150,200
Main model 0,0.76,0.71,0.66,0.62,0.59
Main model 1,0.75,0.68,0.59,0.50,0.45
Main model 2,0.75,0.68,0.59,0.52,0.46
Main model 3,0.76,0.70,0.65,0.59,0.52
Main model 4,0.71,0.63,0.54,0.47,0.44
No persistent  0,0.53,0.42,0.29,0.21,0.14
No persistent  1,0.20,0.10,0.06,0.04,0.03
No persistent  2,0.00,0.00,0.00,0.00,0.00
No persistent  3,0.08,0.06,0.03,0.02,0.02
No persistent  4,0.06,0.03,0.02,0.01,0.01
No random  0,0.71,0.62,0.48,0.38,0.28
No random  1,0.74,0.68,0.60,0.52,0.44
No random  2,0.74,0.65,0.54,0.43,0.34
No random  3,0.74,0.68,0.56,0.45,0.34
No random  4,0.74,0.67,0.58,0.50,0.43
\end{filecontents*}
\csvreader[
			late after line = \\
			]{ablations.csv}%
			{name=\name, 25=\gt, 50=\gs, 100=\gm, 150=\gl, 200=\gh}{%
				\name & \gt & \gs & \gm & \gl & \gh
			}%
			\hline
		\end{tabular}
	\end{table}
	
	\subsection{Additional pseudocodes} \label{sectionAdditionalPseudocode}
	This \lcnamecref{sectionAdditionalPseudocode} gives pseudocodes for lest interesting procedures.
	
	\begin{algorithm}[hbt]
		\caption{RepresentWalk} \label{algorithmRepresentWalk}
		\DontPrintSemicolon
		\KwIn{$G$ - graph with $n$ nodes; \linebreak $\mathcal{W} = v_1 v_2 \ldots v_k$ - a walk on $G$.}
		\KwOut{$\mathbf{x} \in \R^{(n, 3)}$ - the encoded walk.}
		$\mathbf{x} = 0$ \\
		$\mathbf{x}[v_1, 0] = 1$ \\
		$\mathbf{x}[v_k, 1] = 1$ \\
		\For{$i = 1, 2 \ldots k$}{
			$\mathbf{x}[v_i, 2] = 1$
		}
		\Return $\mathbf{x}$
	\end{algorithm}
	
	\begin{algorithm}[tbh]
		\caption{Least degree first heuristic} \label{algorithmLeastDegreeFirst}
		\DontPrintSemicolon
		\KwIn{$G$ - graph with $n$ nodes}
		\KwOut{$\mathcal{W}$ - predicted Hamiltonian cycle}
		$\textnormal{solution} = [~]$  \\
		\For{$i = 1, 2, \ldots n$}{
			\If{$\textnormal{\texttt{Degree}}(i) = \textnormal{\texttt{MaxDegree}}(G)$}{
				$\mathcal{W} = [~]$ \\
				\textnormal{ld} = 0 \hspace{1cm} \mypseudocodecomment{"least degree"}
				\While{$\textnormal{ld} \neq -1$}{
					$\textnormal{ld} = \Argmin_{j \sim \mathcal{W}[-1], j \notin \mathcal{W}} (\texttt{Degree}(j))$ %\tcp*{Returns $-1$ if argument set is empty}
					$\mathcal{W} = \texttt{Append} \left(\mathcal{W}, \textnormal{ld} \right)$
				}
				$\mathcal{W} = \mathcal{W}[:-1]$ \\
				\If{$\textnormal{\texttt{Len}}(\mathcal{W}) > \textnormal{\texttt{\textnormal{solution}}}$}{
					$\textnormal{solution} = \mathcal{W}$
				}
			}
		}	
		\Return solution
	\end{algorithm}
	\bibliographystyle{plain}
	\bibliography{references}